\documentclass{llncs}

\usepackage{amssymb}
\usepackage{color}
\usepackage{graphics}
\usepackage{latexsym}
\usepackage{algorithm}
\usepackage{amsfonts}
\usepackage{amsmath}
\usepackage{float}
\usepackage{tabularx}
\usepackage{subcaption}
\usepackage{hyperref}
\usepackage{graphicx}
\usepackage{float}
\usepackage{tikz}
\usepackage{bm}
\usepackage{tikz-qtree}
\usetikzlibrary{patterns}
\usetikzlibrary{positioning}
\usetikzlibrary{arrows,automata,calc,backgrounds,fit,shapes,shapes.gates.logic.US,shapes.gates.logic.IEC,calc,shapes.misc,circuits.logic.US}
\usepackage{centernot}
\usetikzlibrary{shapes.misc}
\usepackage{pifont}
\usepackage{array}
\usepackage{stmaryrd}
\usepackage{tikz}
\usepackage{xcolor}
\usetikzlibrary[arrows]
\usepackage{multirow}
\usepackage{caption}
\usepackage{adjustbox}

\newcommand{\vars}{\textit{Vars}}
\newcommand{\values}{\textit{Val}}

\newcommand{\mods}[2]{[#2]_{#1}}
\newcommand{\dn}[1]{D\llbracket #1 \rrbracket}
\newcommand{\Rank}[1]{\hspace{3pt} \pmb{\langle} #1 \pmb{\rangle}\hspace{3pt} } 
\newcommand{\areth}{\lozenge}
\newcommand{\comp}{\blacklozenge}
\newcommand{\States}{\Omega}
\newcommand*{\xMin}{0}%
\newcommand*{\yMin}{0}%
\colorlet{lightgray}{gray!10}

\newfloat{program}{thp}{lop}
\floatname{program}{Program}

\begin{document}

\title{RankPL: A Qualitative Probabilistic Programming Language}

\author{Tjitze Rienstra}
\institute{University of Luxembourg\\ \email{tjitze@gmail.com}}

\maketitle

\begin{abstract}
In this paper we introduce \emph{RankPL}, a modeling language
	that can be thought of as a qualitative variant of a probabilistic programming language
	with a semantics based on Spohn's ranking theory.
Broadly speaking, RankPL can be used to represent and reason about processes that exhibit 
	uncertainty expressible by distinguishing ``normal'' from ``surprising'' events.
RankPL allows (iterated) revision of rankings over alternative program states and
	supports various types of reasoning, including abduction and causal inference.
We present the language, its denotational semantics, and a number of practical examples.
We also discuss an implementation of RankPL that is available for download.
\end{abstract}

\section{Introduction}

\emph{Probabilistic programming languages} (PPLs) are programming languages extended with statements to 
	(1) draw values at random from a given probability distribution, and 
	(2) perform conditioning due to observation.
Probabilistic programs yield, instead of a deterministic outcome, a probability distribution over possible outcomes.
PPLs greatly simplify representation of, and reasoning with rich probabilistic models.
Interest in PPLs has increased in recent years,
		mainly in the context of Bayesian machine learning.
Examples of modern PPLs include \emph{Church}, \emph{Venture} and \emph{Figaro}~\cite{DBLP:conf/uai/GoodmanMRBT08,DBLP:journals/corr/MansinghkaSP14,pfeffer2009figaro}, 	
		while early work goes back to Kozen~\cite{DBLP:journals/jcss/Kozen81}.\looseness=-1

\emph{Ranking theory} is a qualitative abstraction of probability theory in which events receive discrete degrees of surprise called \emph{ranks}~\cite{DBLP:books/daglib/0035277}.
That is, events are ranked 0 (not surprising), 1 (surprising), 2 (very surprising), and so on, or $\infty$ if impossible.
Apart from being computationally simpler,
	ranking theory permits meaningful inference without requiring precise probabilities.
Still, it provides analogues to powerful notions known from probability theory, like conditioning and independence.
Ranking theory has been applied in logic-based AI (e.g. belief revision and non-monotonic reasoning~\cite{DBLP:dblp_journals/ai/DarwicheP97,goldszmidt1996qualitative})
		as well as formal epistemology~\cite{DBLP:books/daglib/0035277}.\looseness=-1

In this paper we develop a language called \emph{RankPL}.
Semantically, the language draws a parallel with probabilistic programming in terms of ranking theory.
We start with a minimal imperative programming language (\textbf{if-then-else}, \textbf{while}, etc.)
	and extend it with statements to
	(1) draw choices at random from a given ranking function and
	(2) perform ranking-theoretic conditioning due to observation.
Analogous to probabilistic programs, a RankPL programs yields, instead of a deterministic outcome, a ranking function over possible outcomes.

Broadly speaking, RankPL can be used to represent and reason about processes
	whose input or behavior exhibits uncertainty expressible 
	by distinguishing normal (rank $0$) from surprising (rank $> 0$) events.
Conditioning in RankPL amounts to the (iterated) revision of rankings over alternative program states.
This is a form of revision consistent with the well-known AGM and DP postulates for (iterated) revision~\cite{DBLP:dblp_journals/ai/DarwicheP97,Gardenfors:1995:BR:216136.216138}.
Various types of reasoning can be modeled, including abduction and causal inference.
Like with PPLs, these reasoning tasks can be modeled without having to write inference-specific code.

The overview of this paper is as follows.
Section~\ref{sec:rankingtheory} deals with the basics of ranking theory.
In section~\ref{sec:rpl} we introduce RankPL and present its syntax and formal semantics.
In section~\ref{sec:noisy} we discuss two generalized conditioning schemes (L-conditioning and J-conditioning) and show how they can be implemented in RankPL.
All the above will be demonstrated by practical examples.
In section~\ref{sec:implementation} we discuss our RankPL implementation.
We conclude in section~\ref{sec:conclusion}.

\section{Ranking Theory}\label{sec:rankingtheory}

Here we present the necessary basics of ranking theory, all of which is due to Spohn~\cite{DBLP:books/daglib/0035277}.
The definition of a \emph{ranking function} presupposes a finite set $\Omega$ of possibilities
	and a boolean algebra $\mathcal{A}$ over subsets of $\Omega$, which we call \emph{events}.

\begin{definition}
A ranking function is a function $\kappa: \Omega \rightarrow \mathbb{N} \cup \{\infty\}$ that associates every possibiltiy with a \emph{rank}.
$\kappa$ is extended to a function over events by defining $\kappa(\emptyset) = \infty$ and $\kappa(A) = min( \{ \kappa(w) \mid w \in A \} )$ for each $A \in \mathcal{A} \setminus \emptyset$.
A ranking function must satisfy $\kappa(\Omega) = 0$.
\end{definition}

As mentioned in the introduction, ranks can be understood as degrees of surprise or, alternatively, as inverse degrees of plausibility.
The requirement that $\kappa(\Omega) = 0$ is equivalent to the condition that at least one $w \in \Omega$ receives a rank of 0.
We sometimes work with functions $\lambda: \Omega \rightarrow \mathbb{N} \cup \{\infty\}$ that violate this condition.
The \emph{normalization} of $\lambda$ is a ranking function denoted by $|| \lambda ||$ and defined by $|| \lambda ||(w) = \lambda(w) - \lambda(\Omega)$.
Conditional ranks are defined as follows.

\begin{definition}\label{def:conditional}
Given a ranking function $\kappa$, the rank of $A$ conditional on $B$ (denoted $\kappa(A \mid B)$ is defined by
\[
                \kappa(A \mid B) = \left\{ \begin{array}{ll}
                 \kappa(A \cap B) - \kappa(B)& \mbox{if $\kappa(B) \not = \infty$,} \\
                 \infty & \mbox{otherwise.} \\
                   \end{array}
                  \right. \\
\] 
We denote by $\kappa_{B}$ the ranking function defined by $\kappa_{B}(A) = \kappa(A \mid B)$.
\end{definition}

In words, the effect of conditioning on $B$ is that the rank of $B$ is shifted down to zero (keeping the relative ranks of the possibilities in $B$ constant) 
	while the rank of its complement is shifted up to $\infty$.

How do ranks compare to probabilities?
An important difference is that ranks of events do not add up as probabilities do.
That is, if $A$ and $B$ are disjoint, then $\kappa(A \cup B) = min(\kappa(A), \kappa(B))$, while $P(A \cup B) = P(A) + P(B)$.
This is, however, consistent with the interpretation of ranks as degrees of surprise (i.e., $A \cup B$ is no less surprising than $A$ or $B$).
Furthermore, ranks provide deductively closed beliefs, whereas probabilities do not.
More precisely, if we say that $A$ is believed with firmness $x$ (for some $x > 0$) with respect to $\kappa$ iff $\kappa(\overline A) > x$, 
	then if $A$ and $B$ are believed with firmness $x$ then so is $A \cap B$.
A similar notion of belief does not exist for probabilities, as is demonstrated by the Lottery paradox~\cite{kyburg1961probability}.

Finally, note that $\infty$ and $0$ in ranking theory can be thought of as playing the role of $0$ and $1$ in probability,
	while $min$, $-$ and $+$ play the role, respectively, of $+$, $\div$ and $\times$.
Recall, for example, the definition of conditional probability, and compare it with definition~\ref{def:conditional}.
This correspondence also underlies notions such as (conditional) independence and ranking nets (the ranking-based counterpart of Bayesian networks)
	that have been defined in terms of rankings~\cite{DBLP:books/daglib/0035277}.

\section{RankPL}\label{sec:rpl}

We start with a brief overview of the features of RankPL.
The basis is a minimal imperative language 
	consisting of integer-typed variables, an \textbf{if}-\textbf{then}-\textbf{else} statement and a \textbf{while-do} construct.
We extend it with the two special statements mentioned in the introduction.
We call the first one \emph{ranked choice}.
It has the form
$\{s_1\} \Rank{e} \{s_2\}.$
Intuitively, it states that either $s_1$ or $s_2$ is executed, 
	where the former is a normal (rank 0) event and the latter a typically surprising event whose rank is the value  
	of the expression~$e$.
Put differently, it represents a draw of a statement to be executed, at random, from a ranking function over two choices.
Note that we can set $e$ to zero to represent a draw from two equally likely choices,
	and that larger sets of choices can be represented through nesting.

The second special statement is called the \emph{observe} statement
	$\textbf{observe }b.$
It states that the condition $b$ is observed to hold.
Its semantics corresponds to ranking-theoretic conditioning.
To illustrate, consider the program
	$$\texttt{x} := 10;\mbox{ } \{ \texttt{y} := 1 \} \Rank{1} \{ \{\texttt{y} := 2\} \Rank{1} \{\texttt{y} := 3\} \};\mbox{ } \texttt{x} := \texttt{x} \times \texttt{y};$$
This program has three possible outcomes: $\texttt{x} = 10$, $\texttt{x} = 20$ and $\texttt{x} = 30$, ranked 0, 1 and 2, respectively.
Now suppose we extend the program as follows: 
	$$\texttt{x} := 10;\mbox{ } \{ \texttt{y} := 1 \} \Rank{1} \{ \{\texttt{y} := 2\} \Rank{1} \{\texttt{y} := 3\} \};\mbox{ } \textbf{observe } \texttt{y} > 1;\mbox{ } \texttt{x} := \texttt{x} \times \texttt{y};$$
Here, the observation rules out the event $\texttt{y} = 1$,
	and the ranks of the remaining possibilities are shifted down, 
	resulting in two outcomes $\texttt{x} = 20$ and $\texttt{x} = 30$, ranked 0 and 1, respectively.

A third special construct is the \emph{rank expression} $\textbf{rank }b.$, which evaluates to the rank of the boolean expression $b$.
Its use will be demonstrated later.

\subsection{Syntax}

We fix a set $\vars$ of variables (ranged over by $x$) and denote by $\values$ the set of integers including $\infty$ (ranged over by $n$).
We use $e$, $b$ and $s$ to range over the numerical expressions, boolean expressions, and statements.
They are defined by the following BNF rules:
\noindent \begin{center}
\begin{tabular}{rcl}
$\textit{e}$	& $:=$ 	& $n \mid x \mid \textbf{rank }b \mid (e_1 \hspace{2pt} \areth \hspace{2pt} e_2)$ (for $\areth \in \{ -, +, \times, \div \}$)\\

$\textit{b}$	& $:=$ 	& $\neg b \mid (b_1 \vee b_2) \mid (e_1 \hspace{2pt} \comp \hspace{2pt} e_2)$ (for $\comp \in \{ =, < \}$)\\

$\textit{s}$		& $:=$ 	& $\{s_0; s_1 \} \mid x := e \mid \textbf{if }b\textbf{ then }\{ s_1 \}\textbf{ else } \{ s_2 \} \mid $\\
			& 		& $\textbf{while } b \textbf{ do } \{ s \} \mid \{ s_1 \} \Rank{e} \{ s_2 \} \mid \textbf{observe }b \mid \textbf{skip}$\\
\end{tabular}
\end{center}

We omit parentheses and curly brackets when possible and define $\wedge$ in terms of $\vee$ and $\neg$.
We write $\textbf{if }b\textbf{ then } \{ s \}$ instead of $\textbf{if }b\textbf{ then } \{ s \}$ $\textbf{else }\{\textbf{skip}\}$,
	and abbreviate statements of the form $\{x := e_1\} \Rank{e} \{x := e_2\}$ to $x := e_1 \Rank{e} e_2$.
Note that the \textbf{skip} statement does nothing and is added for technical convenience.

\subsection{Semantics}
\label{sec:formalsemantics}

The denotational semantics of RankPL defines the meaning of a statement $s$ as a function $\dn{s}$ that maps prior rankings into posterior rankings.
The subjects of these rankings are program states represented by \emph{valuations}, i.e., functions that assign values to all variables.
The \emph{initial valuation}, denoted by $\sigma_0$, sets all variables to 0.
The \emph{initial ranking}, denoted by $\kappa_0$, assigns 0 to $\sigma_0$ and $\infty$ to others.
We denote by $\sigma[x \rightarrow n]$ the valuation equivalent to $\sigma$ except for assigning $n$ to $x$.\looseness=-1

From now on we associate $\States$ with the set of valuations and denote the set of rankings over $\States$ by $K$.
Intuitively, if $\kappa(\sigma)$ is the degree of surprise 
	that $\sigma$ is the actual valuation \emph{before} executing $s$,
	then $\dn{s}(\kappa)(\sigma)$ is the degree of surprise 
	that $\sigma$ is the actual valuation \emph{after} executing $s$.
If we refer to the result of running the \emph{program} $s$, we refer to the ranking $\dn{s}(\kappa_0)$.
Because $s$ might not execute successfully, $\dn{s}$ is not a total function over $K$.
There are two issues to deal with.
First of all, non-termination of a loop leads to an undefined outcome.
Therefore $\dn{s}$ is a partial function whose value $\dn{s}(\kappa)$ is defined only if $s$ terminates given $\kappa$.
Secondly, observe statements may rule out all possibilities.
A program whose outcome is empty because of this is said to \emph{fail}.
We denote failure with a special ranking $\kappa_{\infty}$ that assigns $\infty$ to all valuations.
Since $\kappa_{\infty} \not \in K$, we define the range of $\dn{s}$ by $K^{*} = K \cup \{\kappa_{\infty}\}$.
Thus, the semantics of a statement $s$ is defined by a partial function $\dn{s}$ from $K^{*}$ to $K^{*}$.\looseness=-1

But first, we define the semantics of expressions.
A numerical expression is evaluated w.r.t. both a ranking function (to determine values of \textbf{rank} expressions) and a valuation (to determine values of variables).
Boolean expressions may also contain \textbf{rank} expressions and therefore also depend on a ranking function.
Given a valuation $\sigma$ and ranking $\kappa$, we denote by $\sigma_{\kappa}(e)$ the value of the numerical expression $e$ w.r.t. $\sigma$ and $\kappa$,	
	and by $\mods{\kappa}{b}$ the set of valuations satisfying the boolean expression $b$ w.r.t. $\kappa$.
These functions are defined as follows.\looseness=-1\footnote{We omit explicit treatment of undefined operations (i.e. division by zero and some operations involving $\infty$). They lead to program termination.}

\noindent\begin{minipage}{.40\columnwidth}
	\begin{eqnarray*}
	\sigma_\kappa(n)	 				&	=	&	n	\\
	\sigma_\kappa(x)	 				&	=	&	\sigma(x)	\\
	\sigma_\kappa(\textbf{rank }b) 			&	=	&	\kappa(\mods{\kappa}{b}) 	\\
	\sigma_\kappa(a_1 \hspace{2pt} \areth \hspace{2pt} a_2) 			&	=	&	\sigma_\kappa(a_1) \hspace{2pt} \areth \hspace{2pt} \sigma_\kappa(a_2) \\
	\end{eqnarray*}
\end{minipage}
\begin{minipage}{.60\columnwidth}
	\begin{eqnarray*}
	\mods{\kappa}{\neg b}				&	=	&	\States \setminus \mods{\kappa}{b}	\\
	\mods{\kappa}{b_1 \vee b_2} 			&	=	&	\mods{\kappa}{b_1} \cup \mods{\kappa}{b_2}	\\
	\mods{\kappa}{a_1 \hspace{2pt} \comp \hspace{2pt} a_2}			&	=	&	\{ \sigma \in \States \mid \sigma_\kappa(a_1) \hspace{2pt} \comp \hspace{2pt} \sigma_\kappa(a_2) \} \\
	\end{eqnarray*}
\end{minipage}

Given a boolean expression $b$ we will write $\kappa(b)$ as shorthand for $\kappa(\mods{\kappa}{b})$ and $\kappa_{b}$ as shorthand for $\kappa_{\mods{\kappa}{b}}$.
We are now ready to define the semantics of statements. 
It is captured by seven rules, numbered \textbf{D1} to \textbf{D7}.
The first deals with the \textbf{skip} statement, which does nothing and therefore maps to the identity function.
\begin{align}
\dn{\textbf{skip}}(\kappa) 		& = \kappa \tag{\textbf{D1}}. 
\end{align}
The meaning of $s_1; s_2$ is the composition of $\dn{s_1}$ and $\dn{s_2}$.
\begin{align}
\dn{s_1 ; s_2}(\kappa) 		& = \dn{s_2}(\dn{s_1}(\kappa))\tag{\textbf{D2}} 
\end{align}
The rank of a valuation $\sigma$ after executing an assignment $x := e$ 
	is the minimum of all ranks of valuations that equal $\sigma$ after assigning the value of $e$ to $x$.
\begin{align}
\dn{x := e}(\kappa)(\sigma) 	& = \kappa( \{ \sigma' \in \States \mid \sigma = \sigma'[x \rightarrow \sigma'_\kappa(e)] \})\tag{\textbf{D3}} 
\end{align}

To execute $\textbf{if }b\textbf{ then }\{ s_1 \}\textbf{ else }\{ s_2 \}$ we first 
	execute $s_1$ and $s_2$ conditional on $b$ and $\neg b$,
		yielding the rankings $\dn{s_1}(\kappa_{b})$ and $\dn{s_2}(\kappa_{\neg b})$.
	These are adjusted by adding the prior ranks of $b$ and $\neg b$ and
		combined by taking the minimum of the two.
	The result is normalized to account for the case where one branch fails.\looseness=-1
\begin{align*}
\dn{\textbf{if }e\textbf{ then }\{ s_1 \}\textbf{ else }\{ s_2 \}}(\kappa)	& = || \lambda ||,   \\ 
			&  \hspace{-30pt} \mbox{ where }\lambda(\sigma) = min\begin{pmatrix} \dn{s_1}(\kappa_{b})(\sigma) + \kappa(b), \tag{\textbf{D4}} \\ \dn{s_2}(\kappa_{\neg b})(\sigma) + \kappa(\neg b) \end{pmatrix}  
\end{align*}
	Given a prior $\kappa$, the rank of a valuation after executing $s_1 \Rank{e} s_2$
		is the minimum of the ranks assigned by $\dn{s_1}(\kappa)$ and $\dn{s_2}(\kappa)$,
			where the latter is increased by $e$.
	The result is normalized to account for the case where one branch fails.\looseness=-1
\begin{align}
\dn{\{ s_1 \} \Rank{e} \{ s_2 \}}(\kappa)	& = || \lambda ||, \mbox{ where }\lambda(\sigma) = min \begin{pmatrix} \dn{s_1}(\kappa)(\sigma), \\ \dn{s_2}(\kappa)(\sigma) + \sigma_{\kappa}(e) \end{pmatrix} \tag{\textbf{D5}} 
\end{align}
	The semantics of $\textbf{observe }b$ corresponds to conditioning on the set of valuations satisfying $b$, 
		unless the rank of this set is $\infty$ or the prior ranking equals $\kappa_{\infty}$.
\begin{align}
\dn{\textbf{observe }b}(\kappa)  & = 
		\left\{
			\begin{array}{ll}
				\kappa_{\infty}								& \mbox{if }\kappa = \kappa_\infty\mbox{ or }\kappa(b) = \infty\mbox{, or}\\
				\kappa_{b}								& \mbox{otherwise.}\tag{\textbf{D6}}
			\end{array}
		\right. 
\end{align}
	We define the semantics of $\textbf{while }b\textbf{ do }\{s\}$
		as the iterative execution of $\textbf{if }b$ $\textbf{then}$ $\{s\}$ $\textbf{else } \{\textbf{skip}\}$
		until the rank of $b$ is $\infty$ (the loop terminates normally) or the result is undefined ($s$ does not terminate).
	If neither of these conditions is ever met (i.e., if the $\textbf{while}$ statement loops endlessly) then the result is undefined.
\begin{align}
\hspace{-5pt}\dn{\textbf{while }b\textbf{ do }\hspace{-2pt}\{ s \}}(\kappa) &=
		\left\{
			\begin{array}{ll}
				F_{b,s}^{n}(\kappa)	\hspace{4pt}	& \mbox{for the first }n\mbox{ s.t. }F_{b,s}^{n}(\kappa)(b) = \infty \mbox{, or }\\
				\mbox{undef.}		& \mbox{if there is no such }n,
			\end{array}\hspace{-5pt}\tag{\textbf{D7}}
		\right.
\end{align}
where $F_{b,s}: K_{\bot} \rightarrow K_{\bot}$ is defined by $F_{b,s}(\kappa) = \dn{\textbf{if }b\textbf{ then }\{ s \}\textbf{ else } \{ \textbf{skip} \}}(\kappa)$.

Some remarks.
Firstly, the semantics of RankPL can be thought of as a ranking-based variation of the 
	Kozen's semantics of probabistic programs~\cite{DBLP:journals/jcss/Kozen81} (i.e., replacing $\times$ with $+$ and $+$ with $min$).
Secondly, a RankPL implementation does not need to compute complete rankings.
Our implementation discussed in section~\ref{sec:implementation} follows a \emph{most-plausible-first} execution strategy:
	different alternatives are explored in ascending order w.r.t. rank,
	and higher-ranked alternatives need not be explored if knowing the lowest-ranked outcomes is enough, 
	as is often the case.\looseness=-1

\subsubsection{Example}

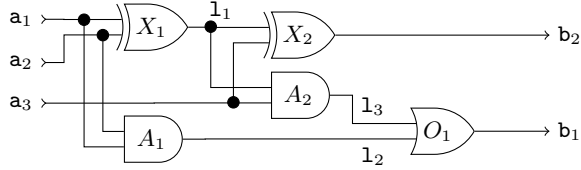
\begin{figure}
\centering
\begin{tikzpicture}[circuit logic US, minimum height=4mm]
\matrix[column sep=10mm,row sep=-1.5mm]
{ \node[yshift=3pt] (i0) {$\texttt{a}_1$};& \node [xor gate] (xor1) {\small$X_1$};	& \node[yshift=-3pt,xor gate] (xor2) {\small$X_2$};	& 	& \node[yshift=-3pt] (o1) {$\texttt{b}_2$}; \\
\node[yshift=2pt] (i1) {$\texttt{a}_2$};&	& \node (cin) {};	&	&  \\
\node[yshift=-3pt] (i2) {$\texttt{a}_3$};&	& \node[and gate] (and1) {\small$A_2$};	&	& \\
	&	\node[and gate,yshift=-3pt] (and2) {\small$A_1$};	&	& \node[or gate] (or1) {\small$O_1$};	& \node (o2) {$\texttt{b}_1$};\\ };
\draw[>-] (i0.east) -- ++(right:3mm) |- (xor1.input 1);
\draw[*-] (xor1.input 1) ++(-5mm,0.8mm) |- (and2.input 2);
\draw[*-] (xor1.input 2) ++(-2.5mm,0.8mm) |- (and2.input 1) ;
\draw[>-] (i1.east) -- ++(right:3mm) |- (xor1.input 2);
\draw (xor1.output) -- ++(right:3mm) |- (xor2.input 1) node[midway,above,xshift=4pt,yshift=0pt] {$\texttt{l}_1$};
\draw[*-] (and1.input 1) ++(-5mm,-2.8mm) |- (xor2.input 2);
\draw[*-] (xor1.output) ++(3mm,0.9mm) |- (and1.input 1);
\draw[>-] (i2.east) -- ++(right:15mm) |- (and1.input 2);
\draw[->] (xor2.output) -- ++(right:3mm) |- (o1.west);
\draw (and1.output) -- ++(right:3mm) |- (or1.input 1) node[midway, above right ] {$\texttt{l}_3$};
\draw (and2.output) -- ++(right:3mm) |- (or1.input 2) node[near end,below right,xshift=+16pt] {$\texttt{l}_2$};
\draw[->] (or1.output) -- (o2.west);
\end{tikzpicture}
\caption{A two-bit full adder}
\label{fig:adder}
\end{figure}

Consider the \emph{two-bit full adder} circuit shown in figure~\ref{fig:adder}.
It contains two \emph{XOR} gates $X_1, X_2$, two \emph{AND} gates $A_1, A_2$ and an \emph{OR} gate $O_1$.
The function of this circuit is to generate a binary representation $(\texttt{b}_1,\texttt{b}_2)$ 
	of the number of inputs among $\texttt{a}_1$, $\texttt{a}_2$, $\texttt{a}_3$ that are high.
The \emph{circuit diagnosis problem} is about explaining observed incorrect behavior by finding minimal sets of gates that, if faulty, cause this behavior.\footnote{See Halpern~\cite[Chapter~9]{DBLP:books/daglib/0014219} for a similar treatment of this example.}

The listing below shows a RankPL solution.
On line 1 we set the constants \texttt{L} and \texttt{H} (representing a \emph{low} and \emph{high} signal); and $\texttt{OK}$ and $\texttt{FAIL}$ (to represent the state of a gate).
Line 2 encodes the space of possible inputs (\texttt{L} or \texttt{H}, equally likely).
The failure variables $\texttt{fa}_1$, $\texttt{fa}_2$, $\texttt{fo}_1$, $\texttt{fx}_2$ and $\texttt{fx}_2$ represent the events of individual gates failing and are set on line 3.
Here, we assume that failure is surprising to degree 1.
The circuit's logic is encoded on lines 4-8, where the output of a failing gate is arbitrarily set to \texttt{L} or \texttt{H}.
Note that $\oplus$ stands for XOR.
At the end we observe $\phi$.

\noindent \begin{center}
\small
\begin{tabular}{rl}
\hline
1 & $\texttt{L} := 0; \texttt{H} := 1; \texttt{OK} := 0; \texttt{FAIL} := 1;$ \\
2 & $\texttt{a}_1 := (\texttt{L} \Rank{0} \texttt{H}); \texttt{a}_2 := (\texttt{L} \Rank{0} \texttt{H}); \texttt{a}_3 := (\texttt{L} \Rank{0} \texttt{H});$ \\
3 & $\texttt{fx}_1 := (\texttt{OK} \Rank{1} \texttt{FAIL}); \texttt{fx}_2 := (\texttt{OK} \Rank{1} \texttt{FAIL}); \texttt{fa}_1 := (\texttt{OK} \Rank{1} \texttt{FAIL}); $ \\
   & \hspace{20pt}$\texttt{fa}_2 := (\texttt{OK} \Rank{1} \texttt{FAIL}); \texttt{fo}_1 := (\texttt{OK} \Rank{1} \texttt{FAIL}); $ \\
4 & $\textbf{if }\texttt{fx}_1 = \texttt{OK}\textbf{ then }\texttt{l}_1 := \texttt{a}_1 \oplus \texttt{a}_2\textbf{ else } \texttt{l}_1 := (\texttt{L} \Rank{0} \texttt{H});$ \\
5 & $\textbf{if }\texttt{fa}_1 = \texttt{OK}\textbf{ then }\texttt{l}_2 := \texttt{a}_1 \wedge \texttt{a}_2\textbf{ else } \texttt{l}_2 := (\texttt{L} \Rank{0} \texttt{H});$ \\
6 & $\textbf{if }\texttt{fa}_2 = \texttt{OK}\textbf{ then }\texttt{l}_3 := \texttt{l}_1 \wedge \texttt{a}_3\textbf{ else } \texttt{l}_3 := (\texttt{L} \Rank{0} \texttt{H});$ \\
7 & $\textbf{if }\texttt{fx}_2 = \texttt{OK}\textbf{ then }\texttt{b}_2 := \texttt{l}_1 \oplus \texttt{a}_3\textbf{ else } \texttt{b}_2 := (\texttt{L} \Rank{0} \texttt{H});$ \\
8 & $\textbf{if }\texttt{fo}_1 = \texttt{OK}\textbf{ then }\texttt{b}_1 := \texttt{l}_3 \vee \texttt{l}_2\textbf{ else } \texttt{b}_1 := (\texttt{L} \Rank{0} \texttt{H});$ \\
9 & $\textbf{observe}\textbf{ }\phi;$\\
\hline
\end{tabular} \end{center} 

The different valuations of the failure variables produced by this program  
	represent explanations for the observation $\phi$, ranked according to plausibility.
Suppose we observe that the input $(\texttt{a}_1,\texttt{a}_2,\texttt{a}_3)$ is valued (\texttt{L},\texttt{L},\texttt{H}) while the output $(\texttt{b}_1,\texttt{b}_2)$ is incorrectly valued (\texttt{H},\texttt{L}) instead of (\texttt{L},\texttt{H}).
Thus, we set $\phi$ to $\texttt{a}_1 = \texttt{L} \wedge \texttt{a}_2 = \texttt{L} \wedge \texttt{a}_3 = \texttt{H} \wedge \texttt{b}_1 = \texttt{H} \wedge \texttt{b}_2 = \texttt{L}$.
The program then produces one outcome ranked 0, namely 
	$(\texttt{fa}_1,\texttt{fa}_2,\texttt{fo}_1,\texttt{fx}_2,\texttt{fx}_2) = (\texttt{OK},\texttt{OK},\texttt{OK},\texttt{FAIL},\texttt{OK})$. 
That is, $\phi$ is most plausibly explained by failure of $X_1$.
Other outcomes are ranked higher than 0 and represent explanations involving more than one faulty gate.\looseness=-1

\section{Noisy Observation and Iterated Revision}\label{sec:noisy}

Conditioning by $A$ means that $A$ becomes believed with infinite firmness.
This is undesirable if we have to deal with iterated belief change or noisy or untrustworthy observations,
	since we cannot, after conditioning on $A$, condition on events inconsistent with $A$.
\emph{J-conditioning}~\cite{goldszmidt1996qualitative} is a more general form of conditioning that addresses this problem.
It is parametrized by a rank $x$ that indicates the firmness with which the evidence must be believed.

\begin{definition}\label{defn:resultoriented}
Let $A \in \Omega$, $\kappa$ a ranking function over $\Omega$ such that $\kappa(A), \kappa(\overline A) < \infty$, and $x$ a rank.
The \emph{J-conditioning} of $\kappa$ by $A$ with strength $x$, denoted by $\kappa_{A \rightarrow x}$, is defined by
	$\kappa_{A \rightarrow x}(B) = min ( \kappa(B | A), \kappa(B | \overline A) + x ).$ 
\end{definition}

In words, the effect of J-conditioning by $A$ with strength $x$ is that $A$ becomes believed with firmness $x$.
This permits iterated belief change, because the rank of $\overline A$ is shifted up only by a finite number of ranks
	and hence can be shifted down again afterwards. 
Instead of introducing a special statement representing J-conditioning, we show that we can already express it in RankPL, 
	using ranked choice and observation as basic building blocks.
Below we write $\kappa_{b \rightarrow x}$ as shorthand for $\kappa_{\mods{\kappa}{b} \rightarrow x}$.
Proofs are omitted due to space constraints.
\begin{theorem}
Let $b$ be a boolean expression and $\kappa$ a ranking function s.t. $\kappa(b) < \infty$ and $\kappa(\neg b) < \infty$.
We then have
$\kappa_{b \rightarrow x} = \dn{\{\emph{\textbf{observe }}b \}\Rank{x} \{\emph{\textbf{observe }}\neg b}(\kappa)\}.$ 
\end{theorem}
\emph{L-conditioning}~\cite{goldszmidt1996qualitative} 
is another kind of generalized conditioning.
Here, the parameter $x$ characterizes the `impact' of the evidence. 
\begin{definition}\label{defn:evidenceoriented}
Let $A \in \Omega$, $\kappa$ a ranking function over $\Omega$ such that $\kappa(A), \kappa(\overline A) < \infty$, and $x$ a rank.
The \emph{L-conditioning} of $\kappa$ is denoted by $\kappa_{A \uparrow x}$ and is defined by
	$\kappa_{A \uparrow x}(B) = min ( \kappa(A \cap B) - y, \kappa(\neg A \cap B) + x - y ),$ 
where $y = min(\kappa(A), x).$
\end{definition}
Thus, L-conditioning by $A$ with strength $x$ means that $A$ improves by $x$ ranks w.r.t. the rank of $\neg A$.
Unlike J-conditioning, L-conditioning satisfies two properties 
that are desirable for modeling noisy observation:
	\textit{reversibility} ($(\kappa_{A \uparrow x})_{\overline{A} \uparrow x} = \kappa$) and 
	\textit{commutativity} ($(\kappa_{A \uparrow x})_{B \uparrow x} = (\kappa_{B \uparrow x})_{A \uparrow x}$)~\cite{DBLP:books/daglib/0035277}.
We can expression L-conditioning in RankPL using ranked choice, observation and the rank expression as basic building blocks.
Like before, we write $\kappa_{b \uparrow x}$ to denote $\kappa_{\mods{\kappa}{b} \uparrow x}$.
\begin{theorem}\label{thm:evidenceoriented}
Let $b$ be a boolean expression, $\kappa$ a ranking function over $\States$ such that $\kappa(b), \kappa(\neg b) < \infty$, and $x$ a rank.
We then have:
	$$\kappa_{b \uparrow x} = D
	\left\llbracket\begin{tabular}{c} 
	\mbox{$\emph{\textbf{if }}(\emph{\textbf{rank}}(b) \leq x)\emph{\textbf{ then}}$} \\ 
	\mbox{$\{\emph{\textbf{observe }}b \} \Rank{x - \emph{\textbf{rank}}(b) + \emph{\textbf{rank}}(\neg b)} \{ \emph{\textbf{observe }}\neg b\}$} \\
	\mbox{$\emph{\textbf{else}}$} \\
	\mbox{$\{\emph{\textbf{observe }}\neg b\} \Rank{\emph{\textbf{rank}}(b) - x} \{\emph{\textbf{observe }}b\}$} \\
	\end{tabular}\right\rrbracket(\kappa)$$
\end{theorem}
In what follows we use the statement $\textbf{observe}_{\textbf{L}}(x, b)$ as shorthand for the statement that represents L-conditioning as defined in theorem~\ref{thm:evidenceoriented}.

\textbf{Example} This example involves both iterated revision and noisy observation.
A robot navigates a grid world and has to determine its location using a map and two distance sensors.
Figure~\ref{fig:localization} depicts the map that we use.
Gray cells represent walls and other cells are empty (ignoring, for now, the red cells and dots).
The sensors (oriented north and south) measure the distance to the nearest wall or obstacle.
To complicate matters, the sensor readings might not match the map.
For example, the $X$ in figure~\ref{fig:localization} marks an obstacle that affects sensor readings, but as far as the robot knows, this cell is empty.\looseness=-1

The listing below shows a RankPL solution.
The program takes as input:
(1) A map, held by an array $\texttt{map}$ of size $\texttt{m}\times \texttt{n}$, 
	storing the content of each cell (0 = empty, 1 = wall);
(2) An array $\texttt{mv}$ (length $\texttt{k}$) of movements (\texttt{N}/\texttt{E}/\texttt{S}/\texttt{W} for north/east/south/west) at given time points; and 
(3) Two arrays $\texttt{ns}$ and $\texttt{ss}$ (length $\texttt{k}$) with distances read by the north and south sensor at given time points.
Note that, semantically, arrays are just indexed variables.\looseness=-1

\noindent \begin{center}
\small
\begin{tabular}{rl}
\hline
Input:	&  $\texttt{k}$: number of steps to simulate\\
Input:	&  $\texttt{mv}$: array (size $\geq \texttt{k}$) of movements (\texttt{N}/\texttt{E}/\texttt{S}/\texttt{W})  \\
Input:	&  \texttt{ns} and \texttt{ss}: arrays (size $\geq \texttt{k}$) of north and south sensor readings \\
Input:	&  $\texttt{map}$: 2D array (size \texttt{m} $\times$ \texttt{n}) encoding the map \\
1	& $\texttt{t} := 0; \texttt{ x} := 0 \Rank{0} \{1 \Rank{0} \{ 2 \Rank{0} \ldots \texttt{m}\} \}; \texttt{ y} := 0 \Rank{0} \{1 \Rank{0} \{ 2 \Rank{0} \ldots \texttt{n}\} \};$ \\
2	& $\textbf{while }(\texttt{t} < \texttt{k})\textbf{ do } \{$ \\
3	& \hspace{20pt}	$\textbf{if }(\texttt{mv}[\texttt{t}] = \texttt{N})\textbf{ then } \texttt{y} := \texttt{y} + 1$ \\
4	& \hspace{20pt}	$\textbf{else if }(\texttt{mv}[\texttt{t}] = \texttt{S})\textbf{ then } \texttt{y} := \texttt{y} - 1$ \\
5	& \hspace{20pt}	$\textbf{else if }(\texttt{mv}[\texttt{t}] = \texttt{W})\textbf{ then } \texttt{x} := \texttt{x} - 1$ \\
6	& \hspace{20pt}	$\textbf{else if }(\texttt{mv}[\texttt{t}] = \texttt{E})\textbf{ then } \texttt{x} := \texttt{x} + 1 \textbf{ else skip;}$ \\
7	& \hspace{20pt}	$\texttt{nd} := 0; \textbf{ while }\texttt{map}[\texttt{x}][\texttt{y} + \texttt{nd} + 1] = 0\textbf{ do }\texttt{nd} := \texttt{nd} + 1;$ \\
8	& \hspace{20pt}	$\textbf{observe}_{\textbf{L}}(1, \texttt{nd} = \texttt{ns}[\texttt{t}]);$ \\
9	& \hspace{20pt}	$\texttt{sd} := 0; \textbf{ while }\texttt{map}[\texttt{x}][\texttt{y} - \texttt{sd} - 1] = 0\textbf{ do }\texttt{sd} := \texttt{sd} + 1;$ \\
10	& \hspace{20pt}	$\textbf{observe}_{\textbf{L}}(1, \texttt{sd} = \texttt{ss}[\texttt{t}]);$ \\
11	& \hspace{20pt}	$\texttt{t} := \texttt{t} + 1;$\\
12	& $\}$ \\
\hline
\end{tabular} 
\end{center} 

The program works as follows.
On line 1 the time point $\texttt{t}$ is set to 0 and the robot's location $(\texttt{x}$, $\texttt{y})$ is set to a randomly chosen coordinate (all equally likely) using nested ranked choice statements.
Inside the \textbf{while} loop, which iterates over $\texttt{t}$,
	we first process the movement $\texttt{mv}[\texttt{t}]$ (lines 3-6).
We then process (lines 7-8) the north sensor reading,
	by counting empty cells between the robot and nearest wall,
	the result of which is observed to equal $\texttt{ns}[\texttt{t}]$---and likewise for the south sensor (lines 9-10).
We use L-conditioning with strength 1 to account for possible incorrect observations.
On line 11 we update $\texttt{t}$.\looseness=-1
\noindent 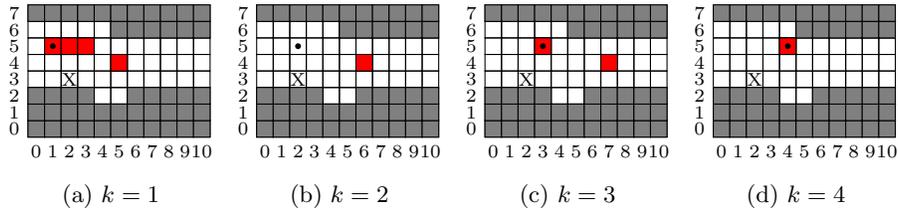
\begin{figure}[t]
\centering
\begin{subfigure}[t]{0.24\columnwidth}
	\label{fig:localizationA}
	\centering
	\begin{tikzpicture}[box/.style={rectangle,draw=black,thin, minimum size=1cm,scale=0.22},scale=0.22]	
	\foreach \x in {0,1,...,10}{
	    \foreach \y in {0,1,...,7}
	        \node[box] at (\x,\y){};
	}
	\foreach \i in {0,...,10} \node[below=3pt] at (\i,\yMin) {\scriptsize$\i$};
	\foreach \i in {0,...,7} \node[left=3pt] at (\xMin,\i) {\scriptsize$\i$};	
\node[box,fill=gray] at (0,0){};\node[box,fill=gray] at (1,0){};\node[box,fill=gray] at (2,0){};\node[box,fill=gray] at (3,0){};\node[box,fill=gray] at (4,0){};\node[box,fill=gray] at (5,0){};\node[box,fill=gray] at (6,0){};\node[box,fill=gray] at (7,0){};\node[box,fill=gray] at (8,0){};\node[box,fill=gray] at (9,0){};\node[box,fill=gray] at (10,0){};\node[box,fill=gray] at (0,1){};\node[box,fill=gray] at (1,1){};\node[box,fill=gray] at (2,1){};\node[box,fill=gray] at (3,1){};\node[box,fill=gray] at (4,1){};\node[box,fill=gray] at (5,1){};\node[box,fill=gray] at (6,1){};\node[box,fill=gray] at (7,1){};\node[box,fill=gray] at (8,1){};\node[box,fill=gray] at (9,1){};\node[box,fill=gray] at (10,1){};\node[box,fill=gray] at (0,2){};\node[box,fill=gray] at (1,2){};\node[box,fill=gray] at (2,2){};\node[box,fill=gray] at (3,2){};\node[box,fill=gray] at (6,2){};\node[box,fill=gray] at (7,2){};\node[box,fill=gray] at (8,2){};\node[box,fill=gray] at (9,2){};\node[box,fill=gray] at (10,2){};\node[box,fill=gray] at (5,6){};\node[box,fill=gray] at (6,6){};\node[box,fill=gray] at (7,6){};\node[box,fill=gray] at (8,6){};\node[box,fill=gray] at (9,6){};\node[box,fill=gray] at (10,6){};\node[box,fill=gray] at (0,7){};\node[box,fill=gray] at (1,7){};\node[box,fill=gray] at (2,7){};\node[box,fill=gray] at (3,7){};\node[box,fill=gray] at (4,7){};\node[box,fill=gray] at (5,7){};\node[box,fill=gray] at (6,7){};\node[box,fill=gray] at (7,7){};\node[box,fill=gray] at (8,7){};\node[box,fill=gray] at (9,7){};\node[box,fill=gray] at (10,7){};Iteration took 1 ms
	\node[font=\scriptsize] at (2,3) {X};
\node[box,fill=red] at (1,5){};\node[box,fill=red] at (5,4){};\node[box,fill=red] at (2,5){};\node[box,fill=red] at (3,5){};
	\fill (1,5) circle[radius=5pt] node {}; 
	\end{tikzpicture}
	\caption{$k = 1$}
\end{subfigure}
\begin{subfigure}[t]{0.24\columnwidth}
	\label{fig:localizationB}
	\centering
	\begin{tikzpicture}[box/.style={rectangle,draw=black,thin, minimum size=1cm,scale=0.22},scale=0.22]	
	\foreach \x in {0,1,...,10}{
	    \foreach \y in {0,1,...,7}
	        \node[box] at (\x,\y){};
	}
	\foreach \i in {0,...,10} \node[below=3pt] at (\i,\yMin) {\scriptsize$\i$};
	\foreach \i in {0,...,7} \node[left=3pt] at (\xMin,\i) {\scriptsize$\i$};
\node[box,fill=gray] at (0,0){};\node[box,fill=gray] at (1,0){};\node[box,fill=gray] at (2,0){};\node[box,fill=gray] at (3,0){};\node[box,fill=gray] at (4,0){};\node[box,fill=gray] at (5,0){};\node[box,fill=gray] at (6,0){};\node[box,fill=gray] at (7,0){};\node[box,fill=gray] at (8,0){};\node[box,fill=gray] at (9,0){};\node[box,fill=gray] at (10,0){};\node[box,fill=gray] at (0,1){};\node[box,fill=gray] at (1,1){};\node[box,fill=gray] at (2,1){};\node[box,fill=gray] at (3,1){};\node[box,fill=gray] at (4,1){};\node[box,fill=gray] at (5,1){};\node[box,fill=gray] at (6,1){};\node[box,fill=gray] at (7,1){};\node[box,fill=gray] at (8,1){};\node[box,fill=gray] at (9,1){};\node[box,fill=gray] at (10,1){};\node[box,fill=gray] at (0,2){};\node[box,fill=gray] at (1,2){};\node[box,fill=gray] at (2,2){};\node[box,fill=gray] at (3,2){};\node[box,fill=gray] at (6,2){};\node[box,fill=gray] at (7,2){};\node[box,fill=gray] at (8,2){};\node[box,fill=gray] at (9,2){};\node[box,fill=gray] at (10,2){};\node[box,fill=gray] at (5,6){};\node[box,fill=gray] at (6,6){};\node[box,fill=gray] at (7,6){};\node[box,fill=gray] at (8,6){};\node[box,fill=gray] at (9,6){};\node[box,fill=gray] at (10,6){};\node[box,fill=gray] at (0,7){};\node[box,fill=gray] at (1,7){};\node[box,fill=gray] at (2,7){};\node[box,fill=gray] at (3,7){};\node[box,fill=gray] at (4,7){};\node[box,fill=gray] at (5,7){};\node[box,fill=gray] at (6,7){};\node[box,fill=gray] at (7,7){};\node[box,fill=gray] at (8,7){};\node[box,fill=gray] at (9,7){};\node[box,fill=gray] at (10,7){};Iteration took 1 ms
	\node[font=\scriptsize] at (2,3) {X};
\node[box,fill=red] at (6,4){};
	\fill (2,5) circle[radius=5pt] node {}; 
	\end{tikzpicture}
	\caption{$k = 2$}
\end{subfigure}
\begin{subfigure}[t]{0.24\columnwidth}
	\label{fig:localizationC}
	\centering
	\begin{tikzpicture}[box/.style={rectangle,draw=black,thin, minimum size=1cm,scale=0.22},scale=0.22]	
	\foreach \x in {0,1,...,10}{
	    \foreach \y in {0,1,...,7}
	        \node[box] at (\x,\y){};
	}
	\foreach \i in {0,...,10} \node[below=3pt] at (\i,\yMin) {\scriptsize$\i$};
	\foreach \i in {0,...,7} \node[left=3pt] at (\xMin,\i) {\scriptsize$\i$};
\node[box,fill=gray] at (0,0){};\node[box,fill=gray] at (1,0){};\node[box,fill=gray] at (2,0){};\node[box,fill=gray] at (3,0){};\node[box,fill=gray] at (4,0){};\node[box,fill=gray] at (5,0){};\node[box,fill=gray] at (6,0){};\node[box,fill=gray] at (7,0){};\node[box,fill=gray] at (8,0){};\node[box,fill=gray] at (9,0){};\node[box,fill=gray] at (10,0){};\node[box,fill=gray] at (0,1){};\node[box,fill=gray] at (1,1){};\node[box,fill=gray] at (2,1){};\node[box,fill=gray] at (3,1){};\node[box,fill=gray] at (4,1){};\node[box,fill=gray] at (5,1){};\node[box,fill=gray] at (6,1){};\node[box,fill=gray] at (7,1){};\node[box,fill=gray] at (8,1){};\node[box,fill=gray] at (9,1){};\node[box,fill=gray] at (10,1){};\node[box,fill=gray] at (0,2){};\node[box,fill=gray] at (1,2){};\node[box,fill=gray] at (2,2){};\node[box,fill=gray] at (3,2){};\node[box,fill=gray] at (6,2){};\node[box,fill=gray] at (7,2){};\node[box,fill=gray] at (8,2){};\node[box,fill=gray] at (9,2){};\node[box,fill=gray] at (10,2){};\node[box,fill=gray] at (5,6){};\node[box,fill=gray] at (6,6){};\node[box,fill=gray] at (7,6){};\node[box,fill=gray] at (8,6){};\node[box,fill=gray] at (9,6){};\node[box,fill=gray] at (10,6){};\node[box,fill=gray] at (0,7){};\node[box,fill=gray] at (1,7){};\node[box,fill=gray] at (2,7){};\node[box,fill=gray] at (3,7){};\node[box,fill=gray] at (4,7){};\node[box,fill=gray] at (5,7){};\node[box,fill=gray] at (6,7){};\node[box,fill=gray] at (7,7){};\node[box,fill=gray] at (8,7){};\node[box,fill=gray] at (9,7){};\node[box,fill=gray] at (10,7){};Iteration took 1 ms
	\node[font=\scriptsize] at (2,3) {X};
\node[box,fill=red] at (3,5){};\node[box,fill=red] at (7,4){};
	\fill (3,5) circle[radius=5pt] node {}; 
	\end{tikzpicture}
	\caption{$k = 3$}
\end{subfigure}
\begin{subfigure}[t]{0.24\columnwidth}
	\label{fig:localizationD}
	\centering
	\begin{tikzpicture}[box/.style={rectangle,draw=black,thin, minimum size=1cm,scale=0.22},scale=0.22]	
	\foreach \x in {0,1,...,10}{
	    \foreach \y in {0,1,...,7}
	        \node[box] at (\x,\y){};
	}
	\foreach \i in {0,...,10} \node[below=3pt] at (\i,\yMin) {\scriptsize$\i$};
	\foreach \i in {0,...,7} \node[left=3pt] at (\xMin,\i) {\scriptsize$\i$};
\node[box,fill=gray] at (0,0){};\node[box,fill=gray] at (1,0){};\node[box,fill=gray] at (2,0){};\node[box,fill=gray] at (3,0){};\node[box,fill=gray] at (4,0){};\node[box,fill=gray] at (5,0){};\node[box,fill=gray] at (6,0){};\node[box,fill=gray] at (7,0){};\node[box,fill=gray] at (8,0){};\node[box,fill=gray] at (9,0){};\node[box,fill=gray] at (10,0){};\node[box,fill=gray] at (0,1){};\node[box,fill=gray] at (1,1){};\node[box,fill=gray] at (2,1){};\node[box,fill=gray] at (3,1){};\node[box,fill=gray] at (4,1){};\node[box,fill=gray] at (5,1){};\node[box,fill=gray] at (6,1){};\node[box,fill=gray] at (7,1){};\node[box,fill=gray] at (8,1){};\node[box,fill=gray] at (9,1){};\node[box,fill=gray] at (10,1){};\node[box,fill=gray] at (0,2){};\node[box,fill=gray] at (1,2){};\node[box,fill=gray] at (2,2){};\node[box,fill=gray] at (3,2){};\node[box,fill=gray] at (6,2){};\node[box,fill=gray] at (7,2){};\node[box,fill=gray] at (8,2){};\node[box,fill=gray] at (9,2){};\node[box,fill=gray] at (10,2){};\node[box,fill=gray] at (5,6){};\node[box,fill=gray] at (6,6){};\node[box,fill=gray] at (7,6){};\node[box,fill=gray] at (8,6){};\node[box,fill=gray] at (9,6){};\node[box,fill=gray] at (10,6){};\node[box,fill=gray] at (0,7){};\node[box,fill=gray] at (1,7){};\node[box,fill=gray] at (2,7){};\node[box,fill=gray] at (3,7){};\node[box,fill=gray] at (4,7){};\node[box,fill=gray] at (5,7){};\node[box,fill=gray] at (6,7){};\node[box,fill=gray] at (7,7){};\node[box,fill=gray] at (8,7){};\node[box,fill=gray] at (9,7){};\node[box,fill=gray] at (10,7){};Iteration took 1 ms
	\node[font=\scriptsize] at (2,3) {X};
\node[box,fill=red] at (4,5){};
	\fill (4,5) circle[radius=5pt] node {}; 
	\end{tikzpicture}
	\caption{$k = 4$}
\end{subfigure}
\caption{Most plausible inferred locations during four iterations}
\label{fig:localization}
\end{figure}

Suppose we want to simulate a movement from $(0,5)$ to $(4, 5)$.
Thus, we use the inputs $\texttt{mv} = \{ E, E, E, E \}$, $\texttt{ns} = \{ 1, 1, 1, 1 \}$, $\texttt{ss} = \{ 2, 1, 2, 3 \}$, 
	while $\texttt{map}$ is set as shown in figure~\ref{fig:localization} (i.e., 1 for every cell containing a wall, 0 for every other cell).
Note that the observed distances stored in $\texttt{ns}$ and $\texttt{ss}$ are consistent with the distances observed along this path,  
	where, at $t = 1$, the south sensor reads a distance of 1 instead of 2, due to the obstacle $X$.

The different values of $\texttt{x},\texttt{y}$ generated by this program encode possible locations of the robot, ranked according to plausibility.
The dots in figure~\ref{fig:localization} show the actual locations, while the red cells represent the inferred most plausible (i.e., rank zero) locations generated by the program.
Terminating after $t = 0$ (i.e., setting $k$ to $1$) yields four locations, all consistent with the observed distances 1 (north) and 2 (south).
If we terminate after $t = 1$, the robot wrongly believes to be at (6, 4), due to having observed the obstacle.
However, if we terminate after $t = 3$, the program produces the actual location.\looseness=-1

Note that using L-conditioning here is essential.
Regular conditioning would cause failure after the third iteration. 
We could also have used J-conditioning, which gives different rankings of intermediate results.

\section{Implementation}\label{sec:implementation}

A RankPL interpreter written in Java can be found at \url{http://github.com/tjitze/RankPL}.
It runs programs written using the syntax described in this paper, 
	or constructed using Java classes that map to this syntax.
The latter makes it possible to embed RankPL programs inside Java code
	and to make it interact with and use classes and methods written Java.
The interpreter is faithful to the semantics described in section~\ref{sec:rpl} and
	implements the \emph{most-plausible-first} execution strategy discussed in section~\ref{sec:formalsemantics}.
All examples discussed in this paper are included, as well as a number of additional examples.

\section{Conclusion and Future Work}\label{sec:conclusion}

We have introduced RankPL, a language semantically similar to probabilistic programming,
	but based on Spohn's ranking theory,
	and demonstrated its utility using examples involving abduction and iterated revision.
We believe that the approach has great potential for applications
	where PPLs are too demanding due to their computational complexity and dependence on precise probability values.
Moreover, we hope that our approach will generate a broader and more practical scope for the topics of 
	ranking theory and belief revision which, in the past, have been studied mostly from purely theoretical perspectives.

A number of aspects were not touched upon and will be addressed in future work.
This includes a more fine grained characterization of termination
	and a discussion of the relationship with nondeterministic programming, 
	which is a special case of RankPL.
Furthermore, we have omitted examples to show that RankPL
	subsumes ranking networks 
	and can be used to reason about causal rules and actions~\cite{goldszmidt1996qualitative}.
We also did not contrast our approach 
	with default reasoning formalisms that use ranking theory as a semantic foundation (see, e.g.,~\cite{DBLP:conf/tark/Pearl90}).

Even though we demonstrated that RankPL is expressive enough to solve 
	fairly complex tasks in a compact manner, it is a very basic language that is best regarded as proof of concept.
In principle, the approach can be applied to any programming language,
	whether object-oriented, functional, or LISP-like.
Doing so would make it possible to reason about ranking-based models expressed using, for example, recursion and complex data structures.
These features are also supported by PPLs such as Church~\cite{DBLP:conf/uai/GoodmanMRBT08}, 
	Venture~\cite{DBLP:journals/corr/MansinghkaSP14} and Figaro~\cite{pfeffer2009figaro}.





\bibliography{bibliography}

\end{document}